\documentclass[conference]{IEEEtran}
\IEEEoverridecommandlockouts
\IEEEoverridecommandlockouts

\usepackage{balance}
 
\IEEEoverridecommandlockouts

\usepackage{cite}
\usepackage{amsmath,amssymb,amsfonts}
\usepackage{soul}

\usepackage{graphicx}
\usepackage{textcomp}
\usepackage{xcolor}
\def\BibTeX{{\rm B\kern-.05em{\sc i\kern-.025em b}\kern-.08em
    T\kern-.1667em\lower.7ex\hbox{E}\kern-.125emX}}

\usepackage{url} 
\usepackage{comment} 
\usepackage{algorithm}
\usepackage{todonotes}

\usepackage[utf8]{inputenc}

\usepackage{siunitx}
\usepackage{array}

\usepackage{makecell}
\usepackage{listings}
\usepackage{float}
\usepackage{balance}
\usepackage[hidelinks]{hyperref}
\usepackage{xspace}
\usepackage{tcolorbox}
\usepackage{enumitem}
\usepackage{multirow}
\usepackage{pifont}
\usepackage{tabularx}
\usepackage[para,online,flushleft]{threeparttable}
\usepackage{colortbl}
\usepackage{subcaption}
\usepackage{adjustbox}
\usepackage{booktabs}
\usepackage[compatibility=false]{caption}
\usepackage{tabularx}
\usepackage{framed}

\newboolean{showcomments}
\setboolean{showcomments}{true}
\ifthenelse{\boolean{showcomments}}
 { \newcommand{\mynote}[2]{
      \fbox{\bfseries\sffamily\scriptsize#1}
        {\small$\blacktriangleright$\textsf{\emph{#2}}$\blacktriangleleft$}}}
        { \newcommand{\mynote}[2]{}}

\begin{document}

\title{evalSmarT: An LLM-Based Framework for Evaluating Smart Contract Generated Comments

} 
\author{
\IEEEauthorblockN{
Fatou Ndiaye Mbodji\textsuperscript{1}, 
Mame Marieme C. Sougoufara\textsuperscript{2}, 
Wendkûuni A. M. Christian Ouedraogo\textsuperscript{1},\\ 
Alioune Diallo\textsuperscript{1},
Kui Liu\textsuperscript{3}, Jacques Klein\textsuperscript{1}, Tegawendé F. Bissyandé\textsuperscript{1}}

\IEEEauthorblockA{
\textsuperscript{1}SnT – University of Luxembourg, 
\textsuperscript{2}Université Cheikh Anta Diop, 
\textsuperscript{3}Huawei
}
}


\maketitle


\begin{abstract}
Smart contract comment generation has gained traction as a means to improve code comprehension and maintainability in blockchain systems. However, evaluating the quality of generated comments remains a challenge. Traditional metrics such as BLEU and ROUGE fail to capture domain-specific nuances, while human evaluation is costly and unscalable. In this paper, we present \texttt{evalSmarT}, a modular and extensible framework that leverages large language models (LLMs) as evaluators. The system supports over 400 evaluator configurations by combining approximately 40 LLMs with 10 prompting strategies. We demonstrate its application in benchmarking comment generation tools and selecting the most informative outputs. Our results show that prompt design significantly impacts alignment with human judgment, and that LLM-based evaluation offers a scalable and semantically rich alternative to existing methods.
\begin{tcolorbox}[colback=gray!5!white, colframe=gray!75!black, title=Resources]
\textit{\underline{Video Demo:}}  
\url{https://youtu.be/HXS_Yiszoz4}  

\vspace{0.3em}
\textit{\underline{Code and Data:}}  
\url{https://anonymous.4open.science/r/SC_code_summarization-4653}  
\end{tcolorbox}
\end{abstract}

\section{Introduction}

Smart contracts represent a foundational technology in decentralized systems, enabling the autonomous execution of agreements. Due to their immutable and financial nature, they are particularly sensitive to design flaws and documentation errors. Clear, accurate, and complete documentation is therefore essential not only for code maintainability, but also for security audits, regulatory compliance, and effective collaboration across teams.

Despite this need, smart contracts are often poorly documented~\cite{pinna2019massive}. Even when comments are present, they frequently lack relevance or alignment with the underlying code. This has motivated the development of automatic comment generation models tailored to smart contracts~\cite{yang2022ccgir, zhao2024automatic}. However, evaluating the quality of these generated comments remains a significant challenge.

Traditional evaluation methods rely on surface-level metrics such as BLEU, METEOR, and ROUGE, which fail to capture the semantic and domain-specific aspects of smart contract behavior. Human evaluation, while more nuanced, is time-consuming, subjective, and difficult to scale. In contrast, the emerging paradigm of using large language models (LLMs) as evaluators referred to as \textit{LLM-as-a-Judge} offers a promising alternative. Prior work has demonstrated the effectiveness of LLMs in evaluating requirements~\cite{10628462} and code summaries~\cite{wu2024can}, particularly when guided by carefully designed prompts.

In this paper, we introduce \texttt{evalSmarT}, a modular and extensible framework for evaluating smart contract comment generation using LLMs. Unlike general-purpose summarization tasks, smart contract documentation requires domain-specific knowledge of Solidity and the Ethereum ecosystem. \texttt{evalSmarT} addresses this challenge by combining multiple LLMs with ten prompting strategies that incorporate domain awareness, language-specific features, and evaluation framing. Our framework enables scalable, reproducible, and semantically rich evaluation of generated comments, bridging the gap between traditional metrics and expert judgment

\section{Motivation and Problem Statement}

Smart contracts are critical components of decentralized systems. Their immutable and financial nature makes them particularly sensitive to design flaws and documentation errors. While automatic comment generation models have emerged to support smart contract comprehension, their evaluation remains limited. Traditional metrics such as BLEU, METEOR, and ROUGE fail to capture domain-specific concerns like security, gas optimization, or Solidity-specific constructs. Human evaluation, though more nuanced, is costly and unscalable.

Below, we present evaluation methods used in existing studies.

\begin{table}[htbp]
\caption{Existing code comment methods and their evaluation metrics}
\begin{center}
\scriptsize
\begin{tabularx}{0.9\linewidth}{|X|X|}
\hline
Method names & Evaluation metrics \\ \hline
SMTranslator \cite{li2020towards} & Human judgment \\ \hline
STAN \cite{li2020stan} & Human judgment \\ \hline
MMTrans \cite{yang2021multi} & BLEU, METEOR, ROUGE \\ \hline
SMARTDOC \cite{hu2021automating} & BLEU, ROUGE, Human judgment \\ \hline
CCGIR \cite{yang2022ccgir} & BLEU, METEOR, ROUGE \\ \hline
SolcTrans \cite{shi2023machine} & BLEU, Human judgment \\ \hline
SCCLLM \cite{zhao2024automatic} & BLEU, ROUGE \\ \hline
SCLA \cite{mao2024scla} & BLEU, METEOR, ROUGE \\ \hline
FMCF \cite{lei2024fmcf} & BLEU, METEOR, ROUGE \\ \hline
SmartBT \cite{xiang2024automating} & BLEU, ROUGE, Human judgment \\ \hline
CCGRA \cite{zhang2023ccgra} & BLEU, METEOR, ROUGE-L, Human judgment \\ \hline
\end{tabularx}
\label{tab:methods_eval}
\end{center}
\end{table}

\begin{tcolorbox}[colback=gray!5, colframe=gray!70, title=Motivation: Lack of Evaluation Diversity]
Most studies rely on traditional automatic metrics such as BLEU, METEOR, and ROUGE, with \textbf{0\% employing LLM-based evaluation} and 55\% incorporating human judgment. This reveals a significant gap in methodological diversity, particularly the absence of modern LLM-based evaluation approaches.
\end{tcolorbox}

\subsection*{Inference of the Diversity Lack}
Although some recent works explore LLMs for the \textbf{generation} of smart contract comments~\cite{xiang2024automating, zhao2024automatic}, to the best of our knowledge, \textbf{none} use LLMs as \textbf{evaluators}. This contrasts with broader blockchain research, where LLMs are increasingly adopted~\cite{he2024large}, and highlights the need to investigate their potential in comment evaluation.

\subsection*{Problem and Objective}
\begin{tcolorbox}[colback=gray!5!white,colframe=gray!75!black,title=Research Problem]
\textbf{Problem Statement:} Despite advances in smart contract comment generation, evaluation practices remain limited to surface-level metrics or manual human judgment. This lack of semantic depth and scalability highlights a critical gap: the potential of large language models (LLMs) as evaluators remains unexplored.

\textbf{Objective:} To design a tool that uses large language models (LLMs) as automated evaluators for smart contract comment generation, aiming to enhance evaluation depth and scalability, while addressing smart contract-specific concerns.
\end{tcolorbox}

\section{System Overview: \texttt{evalSmarT}}

We present \texttt{evalSmarT}, a modular framework for evaluating automatically generated comments for smart contracts. It leverages the \textit{LLM-as-a-Judge} paradigm to assess comment quality. The system is designed to be model-agnostic and can integrate any large language model accessible via local deployment (e.g., through Ollama) or remote APIs (e.g., via OpenRouter). It supports flexible prompt engineering and evaluation workflows, enabling researchers and practitioners to experiment with diverse model–prompt configurations tailored to smart contract documentation.

\texttt{evalSmarT} is implemented using both local (Ollama) and remote (OpenRouter) LLM access, and supports evaluation of comments generated by tools such as SCCLLM, MMTrans, and CCGIR.

\subsection{Evaluation Metrics}
The evaluators assess comment quality across four dimensions: accuracy, completeness, clarity, and helpfulness. The helpfulness metric incorporates audience-specific utility assessment, acknowledging the heterogeneous stakeholder landscape in smart contract ecosystems.

\begin{tcolorbox}[colback=gray!5, colframe=gray!70, title=Evaluation Metrics]
1. Accuracy (0–100)\\
2. Completeness (0–100)\\
3. Clarity (0–100)\\
4. Helpfulness: Identify which audiences would find the comment useful from: \\
\quad - developer\_maintaining\_contract\\
\quad - developer\_reusing\_code\\
\quad - developer\_integrating\_contract\\
\quad - non\_technical\_user\\
\quad - business\_analyst
\end{tcolorbox}

\subsection{Prompting Strategies and Evaluation Protocol}
We define an LLM evaluator as a tuple $\langle M, P \rangle$, where $M$ is the model and $P$ is the prompt template. Prompts are designed to incorporate:
\begin{itemize}
    \item \textbf{Domain knowledge} (e.g., blockchain-specific logic, permission enforcement)
    \item \textbf{Language features} (e.g., Solidity constructs, modifiers, events)
    \item \textbf{Evaluation framing} (e.g., QA-based reasoning)
\end{itemize}

\begin{table}[htbp]
\centering
\caption{Prompting Strategy Design Matrix}
\label{tab:prompt_strategies}
\scriptsize
\begin{tabular}{|l|c|c|c|}
\hline
\textbf{Prompt} & \textbf{Domain} & \textbf{Language} & \textbf{QA} \\
\hline
P1: Baseline & No & No & No \\
P2: Domain-aware & Yes & No & No \\
P3: Language-aware & No & Yes & No \\
P4: Baseline + QA & No & No & Yes \\
P5: Domain + QA & Yes & No & Yes \\
P6: Language + QA & No & Yes & Yes \\
P7: Unguided Domain & Min & No & No \\
P8: Unguided Language & No & Min & No \\
P9: Domain + Language + QA & Yes & Yes & Yes \\
P10: Domain + Language + QA & Yes & Yes & No \\
\hline
\end{tabular}
\vspace{0.2cm}
\footnotesize

\textbf{Legend:} Yes = Full integration; Min = Minimal guidance; No = Not applied
\end{table}

\begin{tcolorbox}[colback=gray!5, colframe=gray!70, title=evalSmart]

\textit{\underline{Type:}} a modular and extensible framework for evaluating automatically generated comments for smart contracts.\\
\textit{\underline{Components:}} Multiple LLM evaluators with diverse prompt strategies;\\
\textit{\underline{Coverage:}} Around  400 evaluators:  40 LLMs with 10 prompting strategies\\
\textit{\underline{Metrics}} accuracy, completeness, clarity, and helpfulness.

\end{tcolorbox}

\section{Demonstration and Use Cases}

We demonstrate \texttt{evalSmarT} through the evaluation of a smart contract comment generation tool. The system loads a set of (code,comment) pairs produced by the tool, applies multiple LLM-based evaluators, and generates structured assessments across four dimensions: accuracy, completeness, clarity, and helpfulness. This process illustrates how \texttt{evalSmarT} can be used to benchmark the performance of comment generation models in a reproducible and scalable manner.

Beyond this core demonstration, \texttt{evalSmarT} enables several practical and research-oriented use cases:

\subsection{Benchmarking for Research}
Researchers can use \texttt{evalSmarT} to compare outputs from multiple comment generation tools. The framework supports up to 400 evaluator configurations (based on approximately 40 LLMs and 10 prompt strategies), allowing for fine-grained analysis of model performance under varied evaluation conditions.

\subsection{Best Output Selection}
In practical settings, \texttt{evalSmarT} ranks multiple generated comments and selects the most appropriate one. This supports developers in choosing the most informative and accurate documentation, especially when integrating or maintaining smart contracts.

\subsection{Prompt and Evaluator Extension}
The system is designed to be extensible. Users can add new prompts or integrate additional models, enabling continuous refinement of evaluation strategies and adaptation to emerging documentation needs in blockchain development.

\begin{tcolorbox}[colback=gray!5!white, colframe=gray!75!black, title=Demonstration Summary]
The demonstration showcases how \texttt{evalSmarT} evaluates the output of a smart contract comment generation tool. A set of code–comment pairs is loaded, and multiple LLM-based evaluators are applied to assess the quality of the generated comments. The system outputs structured scores and justifications across four evaluation dimensions. This demonstration highlights the tool’s ability to benchmark models, select the most informative comment, and support reproducible, scalable evaluation workflows.
\end{tcolorbox}

\section{Illustrative Evaluation and Insights}
To support the demonstration of \texttt{evalSmarT}, we conducted a focused experiment showcasing its evaluation capabilities on real-world smart contract summaries. Rather than presenting an exhaustive benchmark, we aim here to illustrate how the tool operates in practice and to highlight the rationale behind its internal evaluator configuration.

\subsection{Selecting the Default Evaluator}
We experimented with 40 evaluator configurations (10 prompts × 4 LLMs). Among these, the combination of GPT-4 and prompt P6 (language-aware + QA framing) achieved the best alignment with human expert annotations across accuracy, completeness, clarity, and helpfulness dimensions. This configuration serves as the default evaluator in our demonstration.

\subsection{Example: Comparing SCCLLM and CCGIR}

To illustrate the tool in use, we applied \texttt{evalSmarT} to evaluate comments generated by two state-of-the-art smart contract summarization systems: SCCLLM~\cite{zhao2024automatic} and CCGIR~\cite{yang2022ccgir}. We collected real-world smart contract functions from Etherscan, processed them through both tools, and used our selected evaluator (GPT-4 with prompt P6: language-aware + QA) to assess the quality of the generated comments.

The example showcases how \texttt{evalSmarT} facilitates structured comparison between models across the four evaluation dimensions, while also providing audience-specific helpfulness annotations.

\begin{tcolorbox}[colback=gray!5, colframe=gray!70, title=\texttt{evalSmarT} Configuration for Model Comparison]

\textbf{\underline{LLM Component:}} GPT-4 \\
\textbf{\underline{Prompt:}} P6–Language-aware-QA-framing\\
\textbf{\underline{Smart Contract Source:}} Real-world contracts collected from Etherscan\\
\textbf{\underline{Generated Comments:}} Outputs from SCCLLM and CCGIR\\
\textbf{\underline{Expected Output:}} Structured evaluation across accuracy, completeness, clarity, and helpfulness, including identification of relevant audiences
\end{tcolorbox}

\subsection{Findings}

The evaluation revealed clear differences in the performance of the two systems. SCCLLM produced more accurate and complete comments, with better alignment to contract semantics and audience needs. In contrast, CCGIR showed notable weaknesses, particularly when evaluated on smart contracts that differed significantly from its training distribution.

\begin{table}[htbp]
\centering
\scriptsize
\caption{Evaluation scores for SCCLLM using the GPT-4 + P6 (Language-aware + QA) evaluator.}
\label{tab:sccllm_eval}
\begin{tabular}{cccccccccc}
\toprule
Acc. & Comp. & Clar. & Overall & Mnt. & Reuse & Integr. & NonTech & Analyst \\
\midrule
88.53 & 73.90 & 96.22 & 86.22 & 0.97 & 0.99 & 0.76 & 0.02 & 0.06 \\
\bottomrule
\end{tabular}
\end{table}

The results in Table~\ref{tab:sccllm_eval} highlight the strong performance of SCCLLM when evaluated using the GPT-4 + P6 (Language-aware + QA) configuration. The generated comments exhibit high scores in clarity (96.22) and accuracy (88.53), with a slightly lower but still solid score in completeness (73.90). These scores contribute to a robust overall evaluation average of 86.22.

From an audience-specific perspective, the comments are especially helpful for developers maintaining (0.97) or reusing (0.99) the contract, and to a lesser extent for those integrating (0.76) it. However, the helpfulness drops significantly for non-technical users (0.02) and business analysts (0.06). This indicates that while SCCLLM produces highly accurate and readable comments, its utility remains concentrated among technically proficient users. Broader accessibility would likely require additional language simplification or audience-specific tailoring.

\begin{table}[htbp]
\centering
\footnotesize
\setlength{\tabcolsep}{4pt} 
\caption{Evaluation scores for CCGIR using the GPT-4 + P6 (Language-aware + QA) evaluator.}
\label{tab:ccgir_eval}
\begin{tabular}{cccccccccc}
\toprule
Acc. & Comp. & Clar. & Overall & Mnt. & Reuse & Integr. & NonTech & Analyst \\
\midrule
10.00 & 6.00 & 57.00 & 24.33 & 0.40 & 0.40 & 0.40 & 0.00 & 0.00 \\
\bottomrule
\end{tabular}
\end{table}

The evaluation scores for CCGIR (Table~\ref{tab:ccgir_eval}) reveal significantly lower performance compared to SCCLLM. Accuracy (10.00) and completeness (6.00) are particularly low, indicating that the generated comments often fail to correctly and fully describe the smart contract functions. While clarity (57.00) is somewhat better, it remains moderate, suggesting the comments are not sufficiently clear or informative.

Regarding audience-specific helpfulness, CCGIR’s comments show limited utility for developers maintaining, reusing, or integrating the contracts, with only 40

These findings highlight CCGIR’s limitations in generating high-quality and audience-tailored comments for smart contracts, especially when compared to the stronger performance of SCCLLM under the same evaluation conditions.

\begin{tcolorbox}[colback=gray!5, colframe=gray!70, title=Summary of Findings]

\textbf{\underline{SCCLLM~\cite{zhao2024automatic}:}}\\
Generated comments were generally accurate, complete, and clear. Helpfulness tags showed relevance for technical audiences, especially developers reusing or maintaining contracts.

\textbf{\underline{CCGIR~\cite{yang2022ccgir}:}}\\
Comments lacked precision and completeness, often missing key logic. Performance degraded significantly when the input contracts diverged from the types seen during training.

\textbf{\underline{Conclusion:}}\\
\texttt{evalSmarT} revealed that SCCLLM generalizes better to unseen contracts, while CCGIR struggles with out-of-distribution functions. This highlights the importance of evaluation tools that account for generalization, semantic accuracy, and audience relevance.
\end{tcolorbox}

\begin{tcolorbox}[colback=gray!5, colframe=black, title=Tool Summary Snapshot]
\textbf{Users:} Researchers and practitioners working on smart contract comprehension and comment generation.\\
\textbf{Challenge:} Lack of semantic, scalable, and domain-aware evaluation of smart contract comments.\\
\textbf{Method:} LLM-based evaluators using customizable prompt-model configurations for structured evaluation.\\
\textbf{Validation:} Successfully benchmarked SCCLLM and CCGIR; findings align with human judgment and reveal evaluator sensitivity to prompt design.
\end{tcolorbox}

\section{Conclusion}

We presented \texttt{evalSmarT}, a modular and extensible framework for evaluating automatically generated comments for smart contracts. Designed around the \textit{LLM-as-a-Judge} paradigm, the tool enables structured, reproducible, and scalable assessment of comment quality across multiple dimensions. By integrating a wide range of LLMs and prompt strategies, \texttt{evalSmarT} allows users to explore evaluation configurations that align with domain-specific requirements.

This document has focused on demonstrating the tool’s capabilities and design principles. Through a targeted use case involving SCCLLM and CCGIR, we illustrated how \texttt{evalSmarT} supports benchmarking, best-output selection, and flexible evaluator configuration.

The tool is open-source and readily usable by both researchers and practitioners. Future work will include a more comprehensive evaluation campaign and integration of fine-tuned domain-specific evaluators.







\bibliographystyle{IEEEtran}
\bibliography{references}

\end{document}